\pdfoutput=1

\documentclass[11pt]{article}

\usepackage{acl}

\usepackage{times}
\usepackage{latexsym}

\usepackage[T1]{fontenc}

\usepackage[utf8]{inputenc}

\usepackage{microtype}

\usepackage{booktabs}
\usepackage{multirow}
\usepackage{colortbl}
\usepackage{graphicx}
\usepackage{tabularx}
\usepackage{amsmath}
\usepackage{amssymb}
\usepackage{marvosym}
\usepackage{subfigure}
\usepackage{arydshln}
\usepackage{enumitem}
\usepackage{algorithm}
\usepackage{algorithmicx}
\usepackage{algpseudocode}
\usepackage{color}
\usepackage{setspace}
\usepackage[symbol]{footmisc}

\usepackage{tcolorbox}

\definecolor{color1}{rgb}{0.1,0.7,0.8}
\definecolor{color2}{rgb}{0.9,0.1,0.1}
\definecolor{color3}{rgb}{0.7,0.3,0.7}
\definecolor{color4}{rgb}{0.3,0.3,0.7}
\definecolor{color5}{RGB}{8, 102, 3}
\definecolor{color6}{rgb}{0.53, 0.66, 0.42}

\title{Dynamic Data Mixing Maximizes Instruction Tuning \\ for Mixture-of-Experts}

\author{
Tong Zhu$^{1}$\thanks{$\quad$Work was done during an internship at Shanghai AI Laboratory.} , Daize Dong$^{2}$, Xiaoye Qu$^{2}$, Jiacheng Ruan$^{3}$, \\
\textbf{Wenliang Chen}$^{1\text{\Letter}}$, \textbf{Yu Cheng}$^{4\text{\Letter}}$ \\
$^{1}$ Soochow University $^{2}$ Shanghai AI Laboratory  \\
$^{3}$ Shanghai Jiao Tong University  $^{4}$ The Chinese University of Hong Kong \\
\small\texttt{tzhu7@stu.suda.edu.cn,} \texttt{\{dongdaize,quxiaoye\}@pjlab.org.cn,} \texttt{jackchenruan@sjtu.edu.cn} \\
\small\texttt{wlchen@suda.edu.cn,} \texttt{chengyu@cse.cuhk.edu.hk}
}

\begin{document}
\maketitle

\begin{abstract}

Mixture-of-Experts (MoE) models have shown remarkable capability in instruction tuning, especially when the number of tasks scales.
However, previous methods simply merge all training tasks (e.g. creative writing, coding, and mathematics) and apply fixed sampling weights, without considering the importance of different tasks as the model training state changes. 
In this way, the most helpful data cannot be effectively distinguished, leading to suboptimal model performance.
To reduce the potential redundancies of datasets, we make the first attempt and propose a novel dynamic data mixture for MoE instruction tuning.
Specifically, inspired by MoE's token routing preference, we build dataset-level representations and then capture the subtle differences among datasets.
Finally, we propose to dynamically adjust the sampling weight of datasets by their inter-redundancies, thus maximizing global performance under a limited training budget.
The experimental results on two MoE models demonstrate the effectiveness of our approach on both downstream knowledge \& reasoning tasks and open-ended queries.
Code and models are available at \url{https://github.com/Spico197/MoE-SFT} .

\end{abstract}

\section{Introduction}

Instruction tuning is a pivotal step for Large Language Model (LLM) alignment~\cite{openai_introducing_chatgpt,anthropic_introducing_claude}.
To promote the alignment ability, LLMs are typically fine-tuned on a collection of instruction datasets with multiple tasks~\cite{Zhou2023LIMA,mukherjee2023orca,Ouyang2022TrainingLM,lu2024mitigating}.
However, dense models may be constrained by their fixed model capacities when the number of tasks grows in instruction tuning~\cite{Chung2022ScalingIL}.
Instead, Mixture-of-Experts (MoE) naturally incorporates multiple experts, which expands the model capacity~\cite{shazeer2017outrageously,lepikhin2020gshard}, and assigns relevant tokens to specific experts~\cite{fedus2022switch}. 

\begin{figure}[t]
    \centering
    \resizebox{0.9\columnwidth}{!}{
    \includegraphics{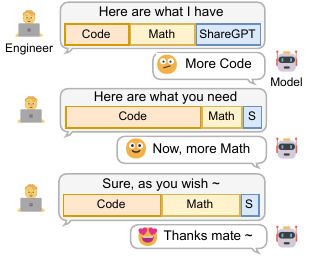}
    }
    \caption{Our proposed dynamic data sampling method for instruction tuning. As the training progresses, the model can dynamically adjust the proportion of data sampling.
    For comparison, previous works concatenate datasets directly and apply fixed sampling weights.
    }
    \label{fig:frontpage}
\end{figure}

To perform instruction tuning, multiple datasets are usually combined in practice~\cite{MosaicML2023MPT}.
In such a complex scenario, datasets from diverse domains may exhibit redundancies, which requires a prudent design in the dataset selection and combination~\cite{Cao2023InstructionMH,Xie2023DoReMiOD}.
Recently, MoE models have demonstrated appealing quality on divergent tasks and reach significantly better performance than dense models, attributed to their excellent task scaling properties~\cite{chen2024textttmoerbench,Shen2023MixtureofExpertsMI}. 
However, how to decide appropriate sampling weights according to models' internal preferences is still under-explored.

Most previous studies~\cite{Shen2023MixtureofExpertsMI,minicpm2024,Wang2023AuroraAC} directly concatenate multiple instruction datasets for supervised fine-tuning (SFT) without considering the sampling weights and task redundancies.
\citet{Jha2023LIMITLI} and \citet{Chen2024LLaVAMoLESM} take sampling weights as a hyper-parameter and find the best combination by handcraft search, which is laborious and costly to enumerate all the combinations.
Thus, it is vital to automatically adjust the sampling weights during the training process with the lowest cost and maximize the alignment abilities.

To this end, we propose a dynamic sampling strategy for MoE models, as illustrated in Figure~\ref{fig:frontpage}.
Our method is based on the hypothesis that \textit{if one dataset is different from the others for the MoE model, there may be fewer redundancies and the sampling weight should be increased in the next round of training}.
Thus, the most important problem is how to identify the differences among datasets considering the model's training state.
It is difficult to build such a meticulous dataset-level difference as the model is constantly changing.
Inspired by the intrinsic properties of MoE models, we formulate the dataset-level representations resorting to specialized experts and token routing preferences~\cite{zoph2022st}. 
Specifically, we count the number of tokens routed to every expert for each dataset, which refers to the gate load.
Afterward, we apply the gate loads as dataset representations and compute L2 distances among them.
Since the distances are obtained from token routing preferences, they could represent the model's internal state.
Finally, we propose a dynamic algorithm to update the sampling weights according to previous sampling weights and current distances.

We experiment on two MoE models with a combination of four representative instruction datasets.
Model performances are evaluated on eight evaluation datasets across knowledge testing, reasoning, and open-ended question answering tasks.
The results demonstrate the effectiveness of our dynamic method.
To help understand the internal mechanism of our method, we also provide thorough analyses of expert specialization and different data combinations.
Our main contributions are summarized as follows:
\begin{itemize}
    \item To our best knowledge, this is the first work to systematically study different sampling methods for MoE models in instruction tuning. Inspired by the inherent attributes of MoE, we introduce a novel dynamic data mixture for combining different instruction datasets. 

    \item 
    To capture the differences among datasets considering the model's training state, we propose to utilize the routing preferences of MoE models to formulate dataset-level representations.

    \item We conduct extensive experiments on two MoE models and validate the effectiveness of our method on a wide range of downstream tasks and open-ended questions.
\end{itemize}

\section{Related Work}

\paragraph{Mixture-of-Experts.}
The Mixture-of-Experts (MoE) is a sparsely activated architecture in neural networks with great efficiency~\cite{shazeer2017outrageously,lepikhin2020gshard,fedus2022switch}.
Attributed to its sparsity, MoE has attracted broad attention in the realm of LLMs~\cite{du2022glam,Jiang2024MixtralOE}.
Subsequent studies follow these model architectures, showing the effectiveness of MoE in dealing with reasoning~\cite{dai2024deepseekmoe}, cross-domain \cite{Li2023SparseMA}, and multi-modal \cite{Mustafa2022MultimodalCL} problems.

\paragraph{Instruction Tuning.}
Instruction tuning is an important step for the LLM alignment.
\citet{Wang2022SelfInstructAL} devise an automatic prompting method to generate enormous instructions and responses with LLMs.
Based on this idea, \citet{Xu2023WizardLMEL} and \citet{Zhao2023APS} further utilize LLMs to generate diverse and complex instructions to enhance the alignment.
Different from the data augmentation methods, \citet{Tunstall2023ZephyrDD} and \citet{Zhou2023LIMA} find a small number of high quality instruction data can boost the alignment performance.
\citet{Cao2023InstructionMH} and \citet{Liu2023DeitaWhatMG} further study data patterns to filter out high quality data to help LLM alignment.
However, none of these approaches consider using different sampling weights when training on multiple instruction datasets.

\paragraph{Dynamic Data Mixing in Pre-training.}
Since there is no relevant literature on dynamic sampling for instruction tuning, we introduce the relevant methods in LLM pre-training.
\citet{Xie2023DoReMiOD} propose DoReMi, a dynamic sampling method for LLM pre-training on multiple domains of data with an extra proxy model for the reference.
\citet{Xia2023ShearedLA} propose to use a series of language models in the same family and estimate the reference loss by fitting scaling law curves.
However, these methods need extra models for estimating reference losses on target domains, which introduces additional training computations.
\citet{Albalak2023EfficientOD} introduce an online data mixing method for LLM pre-training via the multi-armed bandit algorithm.
However, the exploration stage at the beginning of training takes a huge amount of steps, which is not applicable for instruction tuning.
In summary, these dynamic sampling methods are difficult to be transferred into instruction tuning, where the dataset size is relatively small and there are no available proxy models for references.

\section{Preliminaries of Mixture-of-Experts}\label{sec:preliminary}

In a typical MoE structure, the layer is composed of $N$ expert networks $\left\{E_1, E_2, \dots, E_N\right\}$ and a gating network $G$.
Different from common networks, the MoE manifests itself in the design of computational strategy, characterized by inherent sparsity.
Given an input token $x$, the gating network computes a vector of routing scores $G(x)\in\mathbb{R}^N$, denoting the importance of each expert network to process the given input.
The MoE layer then selectively aggregates the outputs from the top-$K$ experts, which is represented as:
\begin{equation}
    y = \sum_{i\in\mathcal{I}_K}{G(x)_i\cdot E_i(x)},
\end{equation}
where $\mathcal{I}_K$ is the set of indices with the highest $K\leq N$ scores in $G(x)$, denoted as:
\begin{equation}
    \mathcal{I}_K = \big\{i_1, \dots, i_K \ | \ G(x)_{i_1} \geq \dots \geq G(x)_{i_N} \big\}.
\end{equation}
To maintain a balanced computational load among experts, an auxiliary balance loss is typically incorporated during the training process.
Given the input dataset $\mathcal{D}_i$, a common practice \cite{shazeer2017outrageously} is to apply a constraint on the routing scores $G(x)$ for each token $x\in\mathcal{D}_i$, which is defined as:
\begin{equation}
    \mathcal{L}_{\mathrm{bal}_i} = \mathrm{CV}(\mathcal{G}_i)^2 + \mathrm{CV}(\mathcal{O}_i)^2,
\end{equation}
where $\mathrm{CV}(\cdot)$ is the function calculating the coefficient of variation from a given vector, measuring the degree of imbalance upon activation.
The $\mathrm{CV}$ score would be high if tokens dispatched to experts are off-balance.
The aggregation of these two terms ensures a balanced dispatching among experts.
The importance score vector $\mathcal{G}_i \in \mathbb{R}^{N}$ corresponds to the summation of routing scores $\sum_{x\in\mathcal{D}_i}{G(x)}$.
The \textbf{gate load vector} $\mathcal{O}_i=\sum_{x\in\mathcal{D}_i}{\mathrm{BinCount}\big(\mathcal{I}_K^{(x)}\big)}, \mathcal{O}_i \in \mathbb{R}^{N}$ is the count of tokens routed to each expert across the entire inputs $\mathcal{D}_i$.
For all the datasets $\mathcal{D}$, we could obtain the gate loads $\mathcal{O} \in \mathbb{R}^{|\mathcal{D}|\times N}$, where $|\mathcal{D}|$ denotes the number of datasets.

\section{Methodology}

In this section, we introduce our dynamic sampling strategy, which automatically adjusts the sampling weights of different instruction datasets. 
After every $m$ steps of model training, we obtain the gate loads $\mathcal{O}$ as dataset-level representations, then calculate the differences across datasets with $\mathcal{O}$ and update sampling weights accordingly.
The dynamic sampling algorithm is presented in Alg~\ref{alg:dynamic-sampling}.

\begin{algorithm}[t]
    \renewcommand{\algorithmicrequire}{\textbf{Input:}}
    \renewcommand{\algorithmicensure}{\textbf{Output:}}
    \caption{\textsc{DynamicSampling}}
    \label{alg:dynamic-sampling}
    \begin{algorithmic}[1]
        \Require sampling weights of last round $\mathbf{w}_{t-1}\in\mathbb{R}^{|\mathcal{D}|}$, normalized gate loads $\hat{\mathcal{O}}\in \mathbb{R}^{|\mathcal{D}|\times N}$, update step size $\eta$, smoothing value $c$, the number of datasets $|\mathcal{D}|$.
        \Ensure updated sampling weights $\mathbf{w}_{t}$.
        \setstretch{1.125} 
        \State \textcolor{gray}{// Update L2 distances across datasets.}
        \State $\delta_{ij} \leftarrow || \hat{\mathcal{O}_i} - \hat{\mathcal{O}_j} ||,\quad\delta\in\mathbb{R}^{|\mathcal{D}| \times|\mathcal{D}|}$
        \State \textcolor{gray}{// Get the average distance for each dataset.}
        \State $\Delta_{i} \leftarrow \left(\sum_j \delta_{ij}\right)\big/\ |\mathcal{D}|,\quad\Delta\in\mathbb{R}^{|\mathcal{D}|}$
        \State \textcolor{gray}{// Calculate the updated sampling weights.}
        \State $\boldsymbol{\alpha} \leftarrow \mathrm{softmax}\left( \log \mathbf{w}_{t-1} + \eta \Delta \right) $
        \State $ \mathbf{w}'_t \leftarrow (1 - c) \boldsymbol{\alpha} + c\ \big/\ |\mathcal{D}|$
        \State \textcolor{gray}{// Normalize sampling weights.}
        \State $\mathbf{w}_t \leftarrow \mathbf{w}'_t\ \big/ \sum\mathbf{w}'_t$
        \State \Return $\mathbf{w}_t$
    \end{algorithmic}
\end{algorithm}

\subsection{Dataset Differences via Gate Load}

As introduced in \S~\ref{sec:preliminary}, the gate load $\mathcal{O}_i\in \mathbb{R}^{N}$ is a vector where each element represents the number of tokens routed to that specific expert.
Since experts in MoE models are well specialized, the token routing distribution can demonstrate the dataset properties.
As discussed in \citet{llama-moe-2023} and \citet{Jiang2024MixtralOE}, deeper layers have better specializations.
Therefore, we calculate the differences among instruction datasets via gate loads in the last layer for each model.

For each dataset $\mathcal{D}_i$, we record the routing tokens and calculate the corresponding gate load $\mathcal{O}_i$.
To alleviate the bias, we discard all padding tokens which may overwhelm the differences across gate loads.
To align the scale of gate loads of different datasets, we normalize $\mathcal{O}_i$ and obtain the final gate load vector $\hat{\mathcal{O}_i} = \mathcal{O}_i / \sum \mathcal{O}$.

After obtaining the gate loads, we calculate the L2 distance $\delta_{ij}$ of each dataset pair $\mathcal{D}_i$ and $\mathcal{D}_j$.
As shown in Line~4 of Alg.~\ref{alg:dynamic-sampling}, we further calculate the averaged distance of one dataset $\mathcal{D}_i$ to all the datasets.
Overall, we obtain $\Delta \in \mathbb{R}^{|\mathcal{D}|}$, which denotes the averaged distance of each dataset.
We further adjust the sampling weights based on $\Delta$.

\subsection{Dynamic Data Sampling}

Based on our hypothesis, if one dataset $\mathcal{D}_i$ is different to the others, the sampling weight of $\mathcal{D}_i$ should be increased since it may contain less redundancies with other datasets.

As presented in Line~6 from Alg. ~\ref{alg:dynamic-sampling}, we calculate the updated sampling weights by adding $\eta\Delta$ to the logarithmic weights of the last time step $\log \mathbf{w}_{t-1}$, where $\eta$ is the update step size that could be regarded as a term similar to the learning rate.
We follow \citet{Xie2023DoReMiOD} and add $c/|\mathcal{D}|$ to smooth and re-normalize the values as shown in Line~7-9 in Alg.~\ref{alg:dynamic-sampling}, where $c$ is a hyper-parameter.

Based on the above strategy, we update the sampling weights every $m$ steps in the training phase.
Following \citet{Xia2023ShearedLA} and \citet{Xie2023DoReMiOD}, the initial sampling weights $\mathbf{w}_0$ is uniformly distributed to alleviate potential biases.

\section{Experiments}

\subsection{Instruction Tuning Datasets}

We use the following four types of instruction datasets for supervised fine-tuning.
In each dataset, we sample 20K instances for training, and 1K instances for gate load evaluation in the sampling weight adjustment.
\textit{(1)} \textbf{ShareGPT.\footnote{\url{https://huggingface.co/datasets/anon8231489123/ShareGPT_Vicuna_unfiltered}}} Multi-turn dialogues with ChatGPT, containing a wide range of open-ended instructions.
\textit{(2)} \textbf{OpenOrca.\footnote{\url{https://huggingface.co/datasets/Open-Orca/OpenOrca}}} Flan~\cite{longpre2023flan} instructions with responses generated by GPT-4 \& GPT-3.5~\cite{OpenOrca}, containing multiple task-oriented instructions.
\textit{(3)} \textbf{Math-Instruct.\footnote{\url{https://huggingface.co/datasets/TIGER-Lab/MathInstruct}}} A collection of math instructions with step-by-step solutions~\cite{yue2023mammoth}.
\textit{(4)} \textbf{Code Instructions.\footnote{\url{https://huggingface.co/datasets/iamtarun/code_instructions_120k_alpaca}}} LLM-generated responses with multiple languages to solve code problems.

\subsection{Evaluation Datasets}

We comprehensively evaluate the ability of models from both Knowledge \& Reasoning (K\&R) and Open-Ended instruction following aspects.
For K\&R, we evaluate the models on MMLU~\cite{hendryckstest2021mmlu}, BigBench-Hard (BBH)~\cite{suzgun2022bbh}, GSM8K~\cite{cobbe2021gsm8k}, MBPP~\cite{austin2021mbpp}, and Question Answering (QA) tasks.
Here, QA consists of ARC-e, ARC-c~\cite{clark2018arc}, and BoolQ~\cite{clark2019boolq}.
Besides, we also report the open-ended instruction following results on MT-Bench.
For more details about evaluation datasets, please refer to Appendix~\ref{sec:eval-datasets-and-metrics}.

\subsection{Baselines}

\noindent\textbf{w/o IT.} The foundation model without instruction tuning.

\noindent\textbf{DataSize. } Static sampling baseline. The sampling weights are determined by the original data size.

\noindent\textbf{Uniform.} Static sampling baseline. The model is fine-tuned with the uniformly distributed sampling weights (all datasets have the same sampling probability).

\noindent\textbf{Random.} A dynamic sampling baseline where sampling weights are assigned with uniformly distributed noise at each round.

\noindent\textbf{Sequential.} Training models on datasets sequentially at each round.

\noindent\textbf{RefLoss.} We use \textbf{Uniform} to estimate the final loss of each dataset as the reference loss, and replace the distance of datasets in Alg~\ref{alg:dynamic-sampling} (line 2) with the loss differences between current loss and reference loss $\Delta_{i} \leftarrow (\mathcal{L}^i_{\text{current}} - \mathcal{L}^i_{\text{reference}})$.
Therefore, \textbf{RefLoss} consumes \textit{2 times of training computation} than the proposed dynamic method.

\subsection{Implementation Details}

We test our method on two MoE models: MoLM 700M-4E (activating 4 experts with 700M parameters)~\cite{shen2023moduleformer} and LLaMA-MoE 3.5B-2E~\cite{llama-moe-2023}. 
We freeze the gate parameters and train models with 2K steps under a global batch size of 128 and a max sequence length of 2048.
The optimizer is AdamW~\cite{Loshchilov2017adamw} with a learning rate of 2e-5, which is warmed up with 3\% steps under cosine scheduling.
Models are trained with gradient checkpointing~\cite{Griewank2000GradientCkpt}, ZeRO-1~\cite{Rajbhandari2019ZeROMO}, and FlashAttention-v2~\cite{dao2023flashattention2}. 
For our proposed dynamic method in LLaMA-MoE, the evaluation interval $m=100$, $\eta$ is 10.0 and $c$ is 5e-2.
In MoLM, $m=200$ and $c$ is 8e-1.
Experiments are conducted on 4$\times$NVIDIA A100 (80G) GPUs.

\subsection{Main Results}

\begin{table*}[t]
\setlength{\tabcolsep}{3.5mm}
\centering
\begin{tabular}{lccccccc}
\toprule
             & \multicolumn{6}{c}{\textbf{Knowledge \& Reasoning}} & \multicolumn{1}{c}{\textbf{Open-Ended}} \\
\multirow{-2}{*}{\textbf{Model}} &
  \textbf{MMLU} &
  \textbf{BBH} &
  \textbf{GSM8K} &
  \textbf{MBPP} &
  \textbf{QA} &
  \textbf{Average} &
  \textbf{MT-Bench} \\
  \cmidrule(lr){1-1} \cmidrule(lr){2-7} \cmidrule(lr){8-8}
\rowcolor[gray]{.93}\multicolumn{8}{c}{\textit{{MoLM}} 700M-4E}                    \\
w/o IT      & 24.73  & \textbf{27.89}  & 1.14   & 5.76   & \textbf{47.52}  & 21.41  & -  \\
DataSize & \textbf{26.62} & 23.94 & \textbf{2.50} & \textbf{10.15} & 43.65 & 21.37 & 2.59 \\
Uniform      & 25.76  & 26.08  & 1.21   & 9.60   & 45.01  & 21.53  & 2.63 \\
\hdashline
Random & 25.95 & 25.94 & 1.59 & 9.49 & \textbf{45.76} & 21.75 & 2.30 \\
Sequential & \underline{26.20} & 26.41 & 1.67 & 9.33 & 45.62 & \underline{21.85} & 2.32 \\
RefLoss & 25.67 & 26.52 & \underline{2.05} & 9.80 & 44.86 & 21.78 & \underline{2.69} \\
\textbf{Dynamic} & 25.83 & \underline{26.96} & 1.82 & \underline{10.12} & 45.28 & \textbf{22.00} & \textbf{2.73} \\
\midrule
\rowcolor[gray]{.93}\multicolumn{8}{c}{\textit{{LLaMA-MoE}} 3.5B-2E}                      \\
w/o IT      & 27.98  & 29.67 & 4.63 & 5.12 & 57.45 & 24.97  & -   \\
DataSize & 31.44 & 29.46 & 1.67 & 11.84 & 59.96 & 26.87 & 4.81 \\
Uniform      & 32.48  & 29.18  & 5.91   & 14.52  & 60.85  & 28.59  & 5.07      \\
\hdashline
Random & \underline{33.39} & 29.43 & 2.73 & \underline{15.80} & \underline{61.17} & 28.50 & 5.00 \\
Sequential & 32.27 & \underline{30.42} & 0.99 & 12.08 & 60.35 & 27.22 & 3.92 \\
RefLoss    & \textbf{33.75}  & 29.02  & \underline{9.63}   & 14.48  & 60.87  & \underline{29.55}  & \underline{5.18}  \\
\textbf{Dynamic}      & 33.07  & \textbf{30.77}  & \textbf{11.90}  & \textbf{16.88}  & \textbf{61.28}  & \textbf{30.78}  & \textbf{5.22} \\
\bottomrule
\end{tabular}
\caption{Main results. Best and the second best results are denoted in \textbf{bold} and \underline{underlined}, respectively.} 
\label{tab:main-results} 
\end{table*}

The main results in Table~\ref{tab:main-results} show that instruction tuning is beneficial for models to enhance their overall abilities on downstream knowledge \& reasoning (K\&R) tasks.
The performance gain from instruction tuning is lower in MoLM than LLaMA-MoE, possibly due to the small model capacity.
For static sampling, the performances of \textbf{DataSize} are lower than \textbf{Uniform}, both in K\&R tasks and open-ended MT-Bench.
Besides, the averaged K\&R score in MoLM \textbf{DataSize} (21.37) is slightly lower than the foundation model (21.41), eliminating the advantage of MoE model's capabilities.

For dynamic sampling, the performances of \textbf{Random} are not stable since it is based on \textbf{Uniform} with random noises.
It achieves better K\&R than \textbf{Uniform} in MoLM, while it is worse in LLaMA-MoE.
\textbf{Sequential} shows the worst MT-Bench scores in both models, demonstrating a bad instruction-following ability.
\textbf{RefLoss} is a strong baseline compared to \textbf{Uniform} and boost the foundation models' performances across the K\&R tasks by 0.37 (MoLM) and 4.58 (LLaMA-MoE).
However, it brings additional training compute due to the reference loss estimation.
Our \textbf{Dynamic} shows great potential and surpasses \textbf{RefLoss} without the additional training cost, which leads to a better and faster convergence.
Overall, \textbf{Dynamic} outperforms other baselines in the averaged K\&R and the MT-Bench results, validating the effectiveness.

\subsection{Analysis}

\subsubsection{Data Combinations}

\begin{figure*}[t]
    \centering
    \resizebox{\textwidth}{!}{
    \includegraphics{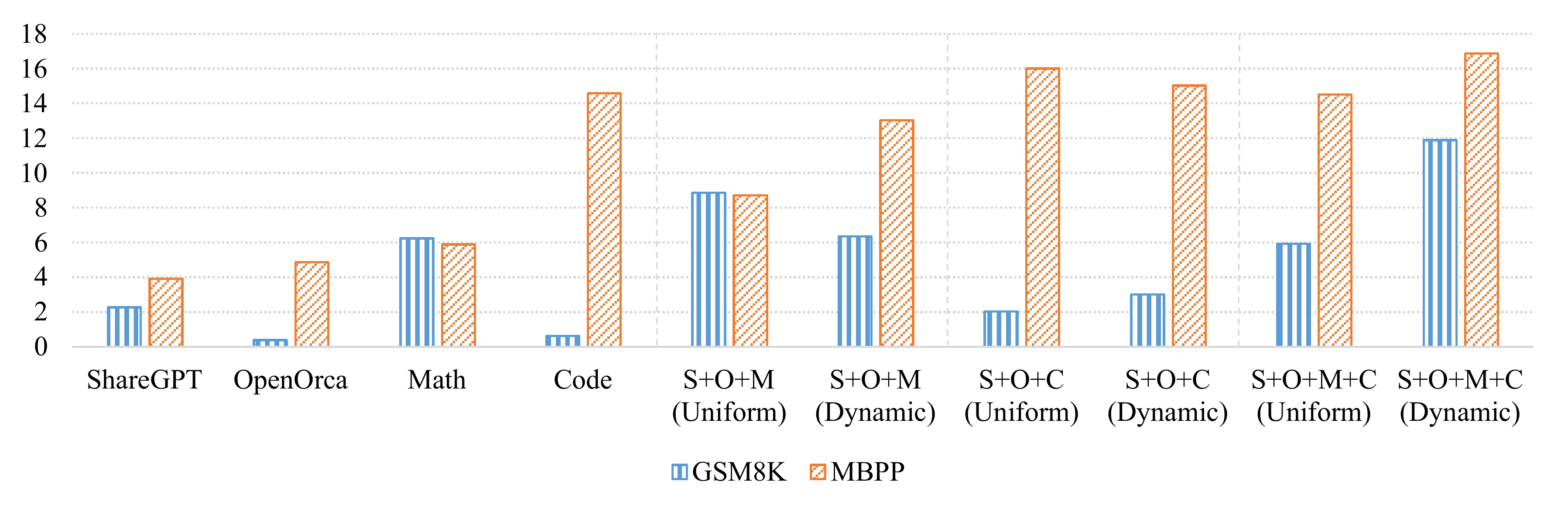}
    }
    \caption{
    Results on different data combinations. LLaMA-MoE 3.5B-2E is fine-tuned for this experiment.
    S, O, M, and C denote for ShareGPT, OpenOrca, Math Instruct, and Code Instructions, respectively.
    }
    \label{fig:data-combination}
\end{figure*}

\begin{figure*}[!t]
\centering
\resizebox{\textwidth}{!}{%
\subfigure[Gate load distances of Uniform]{
    \includegraphics[width=0.31\textwidth]{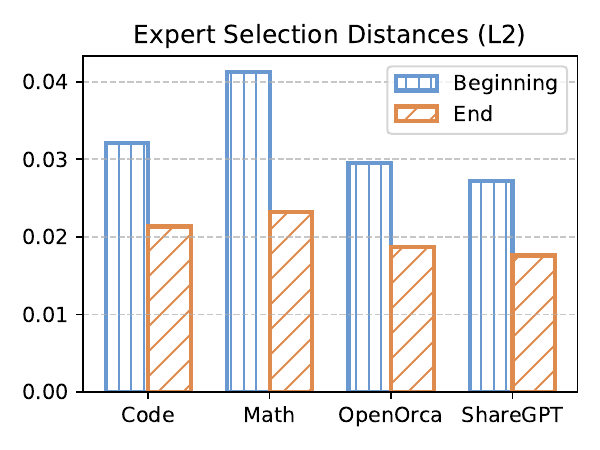}
}
\subfigure[Gate load distances of Dynamic]{
    \includegraphics[width=0.31\textwidth]{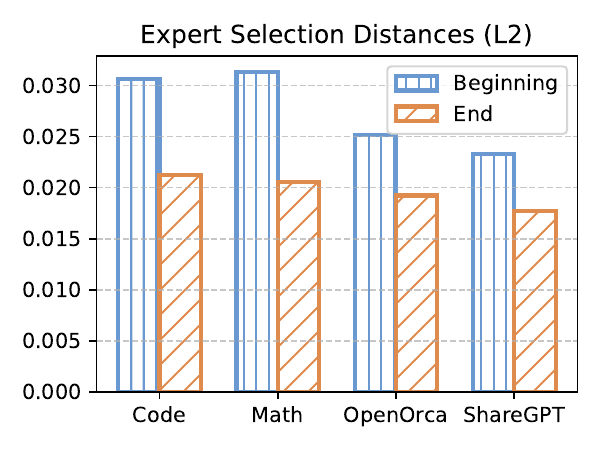}
}
\subfigure[Gate load distances of Dynamic w/o balance loss]{
    \includegraphics[width=0.31\textwidth]{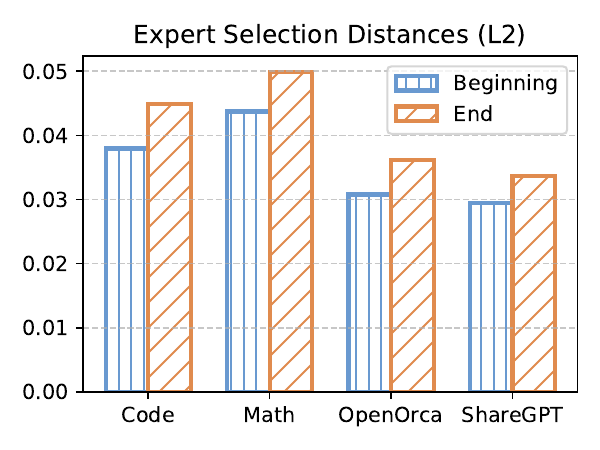}
}
}

\resizebox{\textwidth}{!}{%
\subfigure[$\mathrm{CV}(\mathcal{O}_i)^2$ of Uniform]{
    \includegraphics[width=0.31\textwidth]{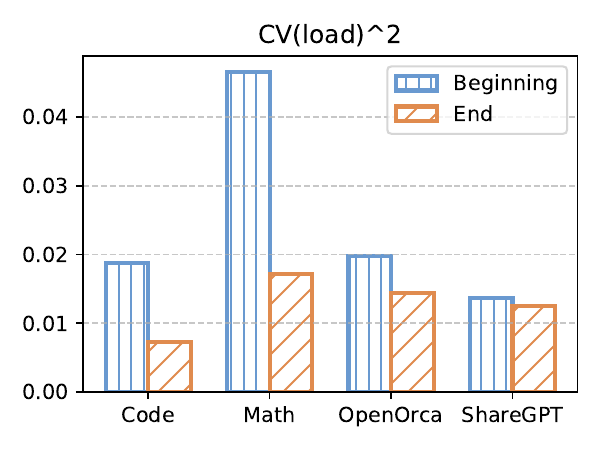}
}
\subfigure[$\mathrm{CV}(\mathcal{O}_i)^2$ of Dynamic]{
    \includegraphics[width=0.31\textwidth]{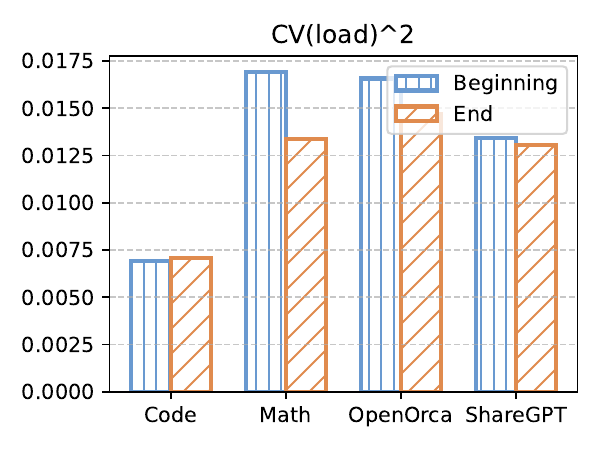}
}
\subfigure[$\mathrm{CV}(\mathcal{O}_i)^2$ of Dynamic w/o balance loss]{
    \includegraphics[width=0.31\textwidth]{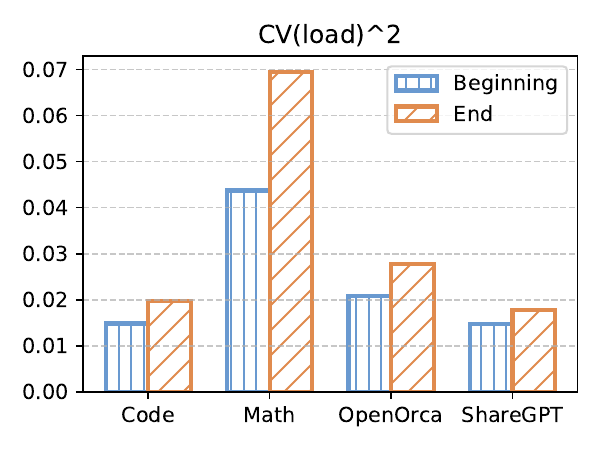}
}
}
\caption{
Gate load differences of LLaMA-MoE 3.5B-2E under different training settings.
If the experts are less specialized after training, the distances and the $\text{CV}(\mathcal{O}_i)^2$ would go down.
For Dynamic and Dynamic w/o balance loss, the ``Beginning'' stands for the first round of evaluation for easier recording.
}
\label{fig:gate-load-diff}
\end{figure*}

\textbf{Q: }\textit{How do datasets contribute to the final performance?}
We conduct experiments on subsets of the training datasets and present the results in Figure~\ref{fig:data-combination}.
Since math and code tasks have strong correlations with the instruction tuning dataset types, we report the GSM8K (math) and MBPP (code) results here.

As shown in the figure, Math-Instruct and Code Instructions are very task-related, and models trained solely on these datasets could reach the best GSM8K and MBPP performances, respectively.
Although the single ShareGPT or OpenOrca is less powerful, it shows great performance when they are combined with Math-Instruct or Code Instruction datasets.
\textbf{Dynamic} is more balanced comparing to the \textbf{Uniform} baseline, where \textbf{Dynamic} strengthens the MBPP performance on math-related combination (S+O+M), and improves the GSM8K performance on code-related combination (S+O+C).
When all four types of datasets are combined for instruction tuning, \textbf{Dynamic} improves both GSM8K and MBPP performances.

\subsubsection{Expert Specialization}

\textbf{Q: }\textit{Does such an gate-load-based dynamic data sampling strategy hurt expert specialization?}
Our method's optimization objective is to make the gate loads more similar across datasets.
Although we freeze the gate parameters during training, the middle activation states may still affect the expert specialization property.
We report the gate load differences and $\mathrm{CV}(\mathcal{O}_i)^2$ for each dataset to measure the expert specialization variations.

As shown in Figure~\ref{fig:gate-load-diff} (abde), we find instruction tuning indeed affects the expert specialization.
However, it is not determined by our gate-load-based distance calculation and dynamic sampling adjustment.
Instead, it is due to the auxiliary balance loss as demonstrated in Figure~\ref{fig:gate-load-diff} (cf).
If we remove the balance loss during training, it would lead to more specialized experts, but the performance would be lower according to Table~\ref{tab:ablation}.

\subsubsection{Evaluation Interval}

\begin{figure*}[t]
\centering
\subfigure[$m=200$]{
    \includegraphics[width=0.23\textwidth]{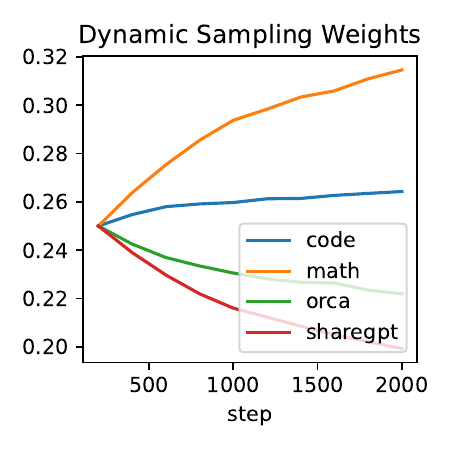}
}
\hfill
\subfigure[$m=100$]{
    \includegraphics[width=0.23\textwidth]{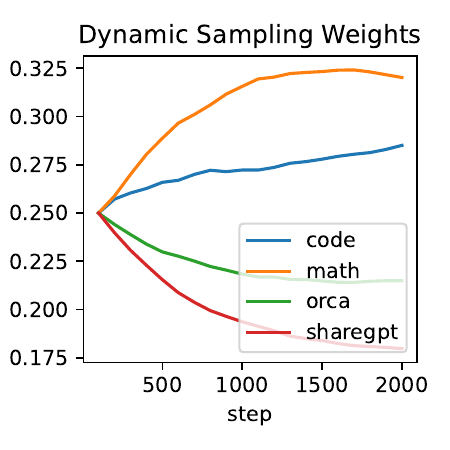}
}
\hfill
\subfigure[$m=50$]{
    \includegraphics[width=0.23\textwidth]{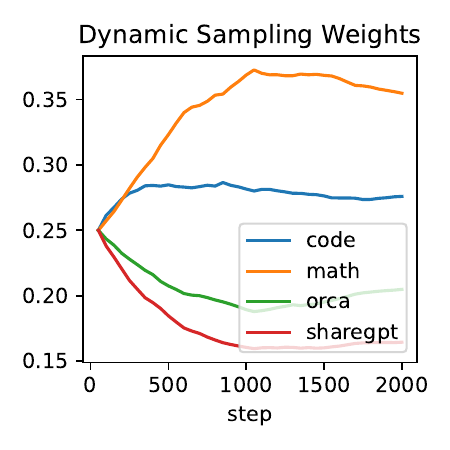}
}
\hfill
\subfigure[$m=20$]{
    \includegraphics[width=0.23\textwidth]{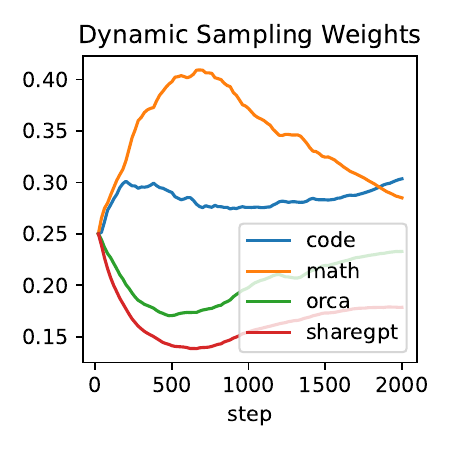}
}
\caption{
Dynamic sampling weights with different evaluation intervals.
Experiments are conducted on LLaMA-MoE 3.5B-2E.
}
\label{fig:eval-interval}
\end{figure*}

\textbf{Q: }\textit{How does the evaluation interval affect the performance?}
Our dynamic sampling weights strategy is applied every $m$ training steps.
Here we investigate the effect of the evaluation intervals by conducting experiments with different $m$ values.

As shown in Figure~\ref{fig:eval-interval}, the evaluation interval is crucial to the sampling weights update and may vary a lot with different $m$ values.
When $m=200$, the sampling weights do not converge and monotonically go up or down.
However, when $m=20$, there are more sampling weights adjustments, leading to training instability as the differences in gate loads may have reversals.
Comparing to the convergence status in Figure~\ref{fig:eval-interval} and results in Table~\ref{tab:evaluation-interval}, we take $m=100$ as the best practice.

\begin{table}[]
    \centering
    \resizebox{0.8\columnwidth}{!}{
    \begin{tabular}{ccc}
    \toprule
    Evaluation Interval & BBH & GSM8K \\
    \midrule
    200 & 29.21 & 8.19 \\
    100 & \textbf{30.77} & \textbf{11.90} \\
    50 & 29.04 & 7.58 \\
    20 & 28.98 & 5.99 \\
    \bottomrule
    \end{tabular}
    }
    \caption{
    LLaMA-MoE 3.5B-2E performances with different evaluation intervals.
    }
    \label{tab:evaluation-interval}
\end{table}

\subsubsection{Learning Efficiency}

\textbf{Q: }\textit{How does the number of training steps affect the results?}
We change the number of training steps and freeze the other hyper-parameters to observe the trend of performance variation.

\begin{figure}[t]
\centering
\subfigure{
    \includegraphics[width=0.22\textwidth]{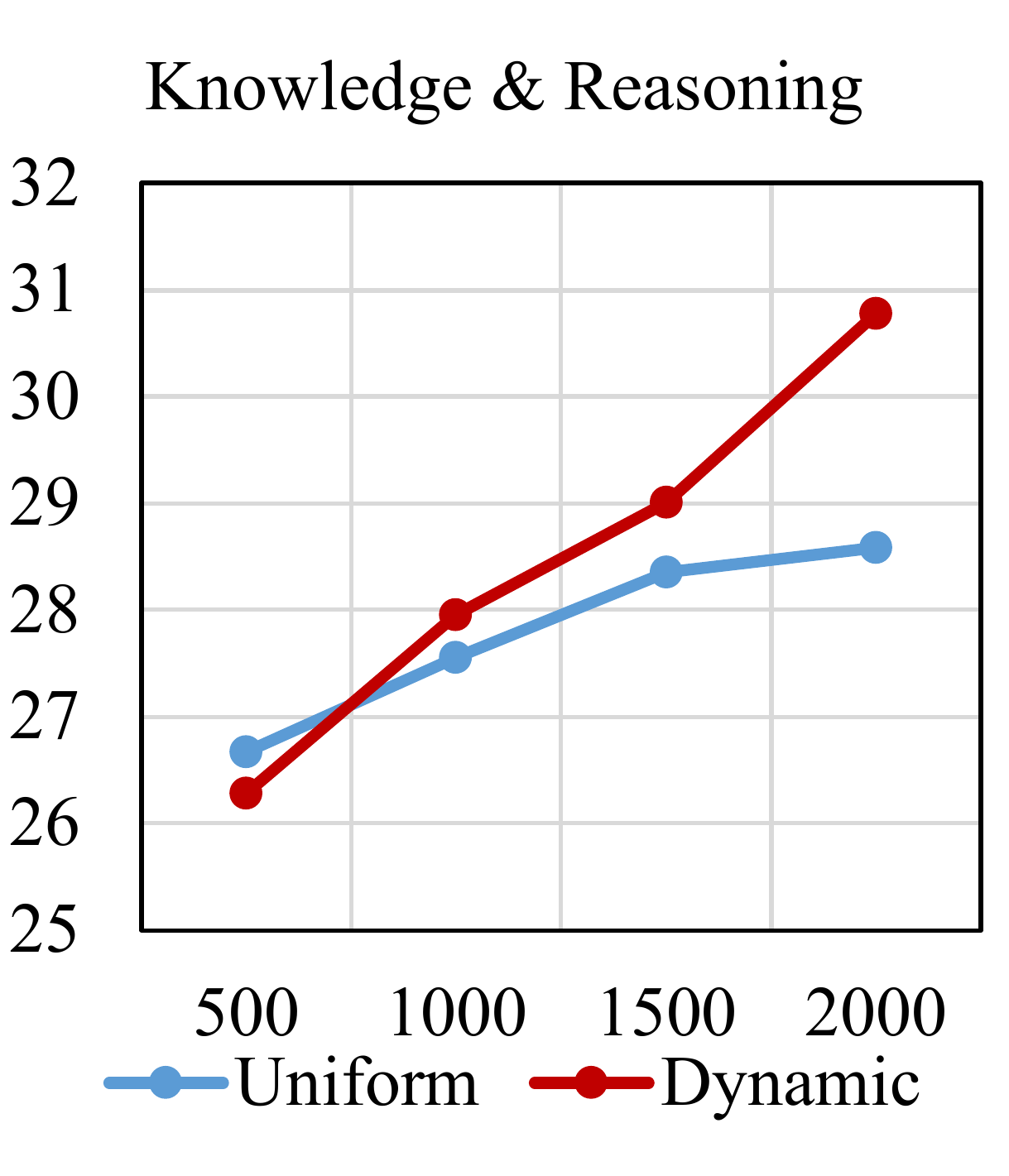}
}
\subfigure{
    \includegraphics[width=0.22\textwidth]{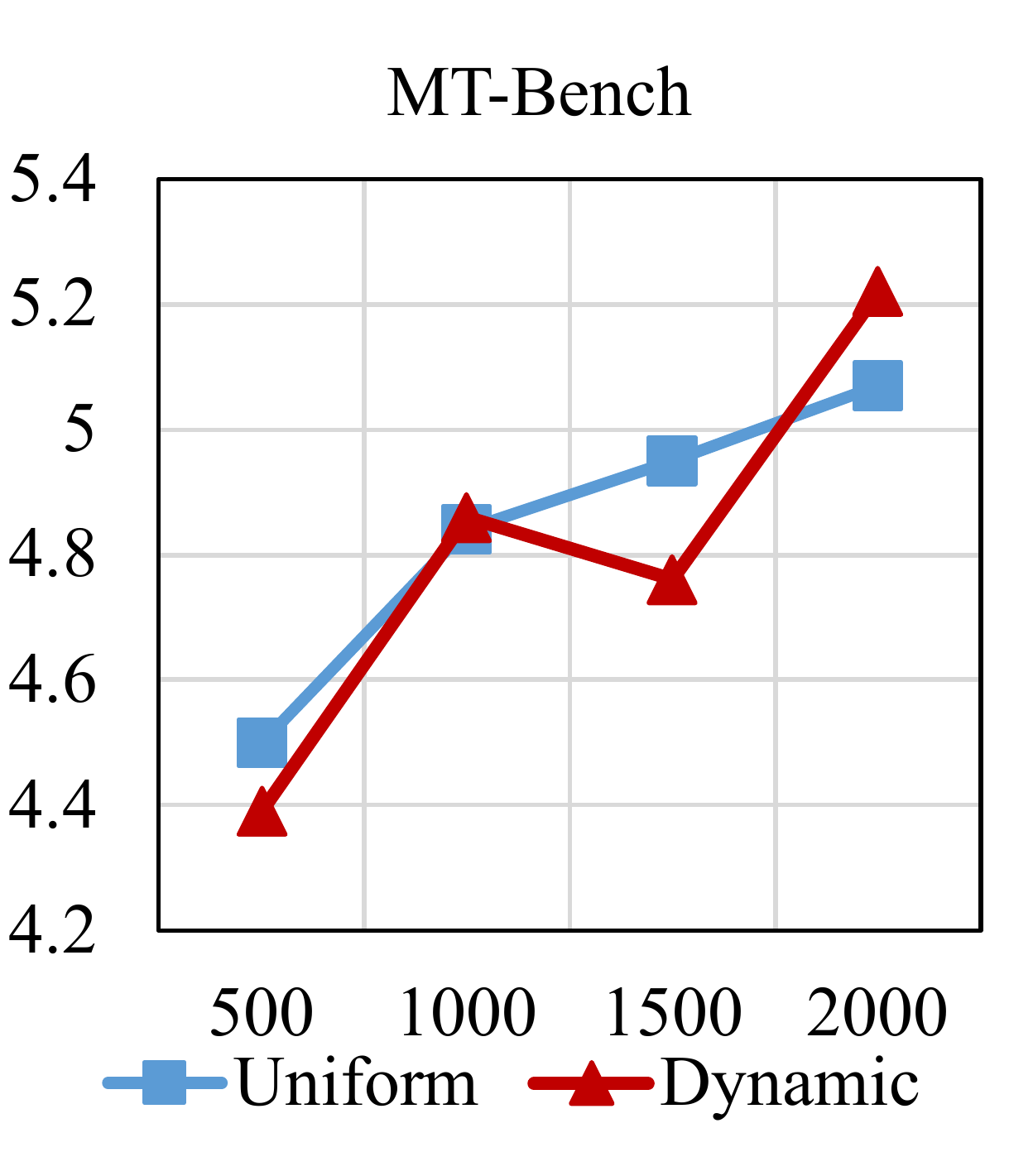}
}
\caption{
Performances with different training steps.
Experiments are conducted on LLaMA-MoE 3.5B-2E.
}
\label{fig:training-step}
\end{figure}

From Figure~\ref{fig:training-step}, both \textbf{Uniform} and \textbf{Dynamic} benefits from more training steps, and they consistently improve the performance on knowledge and reasoning tasks.
Even 500 steps can make the fine-tuned model outperforms the foundation model (Uniform 26.67 \& Dynamic 26.28 vs. w/o IT 24.97).
As the number of training steps grows, \textbf{Uniform} seems to reach its performance ceiling, and the gap between these two methods further increases.
As to the open-ended performance on MT-Bench, the \textbf{Dynamic} method has more fluctuations, but it could outperforms the \textbf{Uniform} baseline as more training steps are applied.

\subsubsection{Other Sampling Weights}

\begin{table*}[t]
\centering
\resizebox{0.9\textwidth}{!}{%
\begin{tabular}{lccccccc}
\toprule
             & \multicolumn{6}{c}{\textbf{Knowledge \& Reasoning}} & \multicolumn{1}{c}{\textbf{Open-Ended}} \\
\multirow{-2}{*}{\textbf{Model}} &
  \textbf{MMLU} &
  \textbf{BBH} &
  \textbf{GSM8K} &
  \textbf{MBPP} &
  \textbf{QA} &
  \textbf{Average} &
  \textbf{MT-Bench} \\
  \cmidrule(lr){1-1} \cmidrule(lr){2-7} \cmidrule(lr){8-8}
w/o IT      & 27.98  & 29.67  & 4.63   & 5.12   & 57.45  & 24.97  & -     \\
\midrule
\rowcolor[gray]{.93}\multicolumn{8}{c}{\textit{Static Sampling}}                      \\
DataSize      & 31.44  & 29.46  &  1.67  &  11.84 &  59.96 & 26.87 & 4.81 \\
Uniform      & 32.48  & 29.18  & 5.91   & 14.52  & 60.85  & 28.59  & 5.07       \\
FinalStatic & 32.84 & 30.11 & 9.93 & 14.61 & 60.93 & 29.68 & 5.11 \\
\midrule
\rowcolor[gray]{.93}\multicolumn{8}{c}{\textit{Static Distances}}                      \\
SentEmb      & 33.85  &  29.70 &  7.66  &  16.29 & 61.75  &  29.85 & 5.21 \\
GateLoad      & 32.75  &  29.98 &  6.60  & 14.07  & 61.78   & 29.04 & 4.98 \\
\midrule
\rowcolor[gray]{.93}\multicolumn{8}{c}{\textit{Initial Sampling Weights}} \\
Dynamic$_{\text{SentEmb}}$      & 33.46 & 29.02 & 8.95 & 15.68 & 61.03 & 29.63 & 5.16 \\
Dynamic$_{\text{Uniform}}$      & 33.07  & 30.77  & 11.90  & 16.88  & 61.28  & 30.78  & 5.22 \\
\bottomrule
\end{tabular}
}%
\caption{Other sampling weights. Experiments are conducted on LLaMA-MoE 3.5B-2E.}
\label{tab:other-sampling-weights}
\end{table*}

\textbf{Q: }\textit{What if we use the final sampling weights obtained from the proposed \textbf{Dynamic} to train the model again?}
To find whether the final sampling weights of \textbf{Dynamic} provide a good data combination for an MoE model, we conduct the experiments on LLaMA-MoE.

As presented in Table~\ref{tab:other-sampling-weights}, \textbf{FinalStatic} is better than \textbf{Uniform} and \textbf{DataSize} in both K\&R tasks and MT-Bench.
Surprisingly, compared to the results in Table~\ref{tab:main-results}, \textbf{FinalStatic} (29.68) is even better than \textbf{RefLoss} (29.55) in the averaged K\&R score.
This indicates that our dynamic method could help find better sampling weights even on static sampling.
In addition, \textbf{FinalStatic} is still worse than \textbf{Dynamic}, which verifies the model's internal state changing.
Thus, dynamic sampling could reach a better performance than static sampling.

\textbf{Q: }\textit{Similar datasets are redundant, how does this hypothesis hold? What if we use sentence embedding to compute the dataset differences instead of gate loads?}
In order to verify the effectiveness of the gate load versus the sentence embedding distances, we conduct utilize SentenceTransformers~\cite{reimers-2019-sentence-bert} to replace the input gate loads $\mathcal{O}$ in Alg.~\ref{alg:dynamic-sampling} and compute L2 distances afterwards.

As shown in Table~\ref{tab:other-sampling-weights}, 
\textbf{SentEmb} outperforms \textbf{Uniform} across the tasks, which indicates the effectiveness of dataset re-weighting by their inter similarities.
The averaged \textbf{GateLoad} performance is lower than \textbf{SentEmb} in both the averaged knowledge \& reasoning tasks and the open-ended MT-Bench.
Nevertheless, \textbf{SentEmb} could not be easily applied to make constant improvements in the whole training phase.
Although \textbf{GateLoad} is worse than \textbf{SentEmb}, the model benefits from the iterative sampling weights adjustments, and \textbf{Dynamic} surpasses \textbf{SentEmb} in both K\&R and open-ended performances.

In addition to further verify the hypothesis, we compare it to the counterpart (similar datasets should have \textit{greater} sampling weights) and present the results in Appendix~\ref{sec:inverse-hypothesis}.

\textbf{Q: }\textit{What about other initial sampling weights rather than the uniform distribution?}
Since \textbf{SentEmb} has better performance than \textbf{Uniform} and \textbf{GateLoad}, we wonder if it is better to apply its sampling weights as the initial ones rather than the uniform distribution.

The results in Table~\ref{tab:other-sampling-weights} show that the uniform initialized \textbf{Dynamic$_{\text{Uniform}}$} outperforms \textbf{Dynamic$_{\text{SentEmb}}$} (30.78 vs. 29.63 in K\&R, 5.22 vs. 5.16 in MT-Bench), which is in line with the conclusions in \citet{llama-moe-2023}.
We conjecture that the imbalanced initial weights would bring biases and make the model hard to convergence.

\subsubsection{Ablation Study}

\begin{table}[]
    \centering
    \begin{tabular}{lcc}
    \toprule
    Model & Avg. K\&R & MT-Bench \\
    \midrule
    LLaMA-MoE & 30.78 & 5.22 \\
    $\quad$w/o frozen gate & 28.78 & 4.91 \\
    $\quad$w/o balance loss & 29.38 & 4.88 \\
    $\quad$w/o gate noise & 30.04 & 4.98 \\
    \bottomrule
    \end{tabular}
    \caption{
    Ablation study.
    Avg. K\&R stands for the averaged score of knowledge \& reasoning tasks (MMLU, BBH, Math, and Code).
    }
    \label{tab:ablation}
\end{table}

There are differences between sparse MoE models and dense models during training due to their specific techniques.
Here we investigate the effectiveness of fronzen gate, balance loss, and gate noise for instruction tuning on MoE.

The results are presented in Table~\ref{tab:ablation}.
Similar to \citet{Shen2023MixtureofExpertsMI}, we find the frozen gate, balance loss, and gate noise have all positive effects to the model performances.
Frozen gate is to freeze the gate networks and the gate projections in FFNs when fine-tuning.
This leads to better performance as the gate is well trained during the pre-training stage, and instruction tuning may break the specialized token routing property.
Balance loss and gate noise are beneficial to model training since they are in line with the pre-training objectives.

\section{Conclusion}

To combine different datasets and maximize the MoE model's alignment ability, we assign different sampling weights to corresponding datasets.
By incorporating the internal model state and the dataset properties, we propose to use the gate load from MoE models to obtain dataset representations.
Based on the representations, we calculate distances between each pair of datasets, indicating the inter-redundancies.
We further devise an automatic algorithm to dynamically update the sampling weights.
The proposed method outperforms other baselines and demonstrate good performance on knowledge \& reasoning tasks and open-ended question answering.

\section*{Limitations}

\paragraph{\textbf{More Models.}} Due to the limit computing resources, we test the method's effectiveness on two representative decoder-style MoE models.
Dynamic sampling on larger models like Mixtral~\cite{Jiang2024MixtralOE} is currently not verified.

\paragraph{\textbf{Number of Datasets.}} For a combination of two datasets, there are no differences between the distance vector $\Delta$, so the dynamic sampling method does not take into effect and the sampling weights would stay unchanged.
Therefore, there should be at least three instruction tuning datasets for applying our method.

\bibliography{anthology,custom}
\bibliographystyle{acl_natbib}

\appendix

\section{Appendix}
\label{sec:appendix}

\subsection{Inverse Hypothesis}\label{sec:inverse-hypothesis}

To validate the proposed hypothesis, we conduct experiments on the counterpart one (denoted as \textbf{Inverse}), where \textit{similar} datasets would have \textit{greater} sampling weights in the next round during training.

\begin{figure}[t]
\centering
\subfigure[Dynamic]{
    \includegraphics[width=0.46\columnwidth]{src/figs/weights-interval-100.pdf}
}
\hfill
\subfigure[Inverse]{
    \includegraphics[width=0.46\columnwidth]{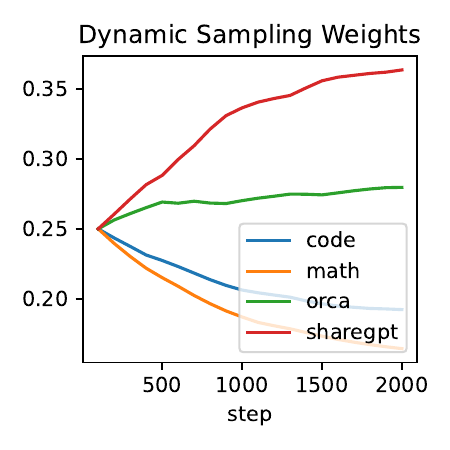}
}
\caption{
Dynamic sampling weights of different hypotheses.
Experiments are conducted on LLaMA-MoE 3.5B-2E.
}
\label{fig:inverse-hypothesis-sampling-weights}
\end{figure}

As illustrated in Figure~\ref{fig:inverse-hypothesis-sampling-weights}, the Inverse sampling method lead to different sampling weights compared to \textbf{Dynamic}.
As shown in Table~\ref{tab:inverse-hypothesis-results}, the performance of \textbf{Inverse} is imbalanced, where GSM8K (5.84 vs. 11.90) is much lower than \textbf{Dynamic}.
The scores of MT-Bench also show that the \textbf{Inverse} method would bring an adverse effect and the performance is even lower than \textbf{Uniform}.

These results demonstrate that our proposed hypothesis is both intuitive and effective.

\begin{table}[t]
    \centering
    \begin{tabular}{cccc}
     \toprule
      Method & GSM8K & MBPP & MT-Bench \\
     \midrule
      Inverse & 5.84 & 17.27 & 4.65 \\
      Uniform & 5.91 & 14.52 & 5.07 \\
      Dynamic & 11.90 & 16.88 & 5.22 \\
     \bottomrule
    \end{tabular}
    \caption{Inverse-hypothesis results of LLaMA-MoE 3.5B-2E, where the sampling weights of similar datasets would be increased in the next round.}
    \label{tab:inverse-hypothesis-results}
\end{table}

\subsection{Datasets and Metrics for Evaluations}\label{sec:eval-datasets-and-metrics}
Here we introduce the datasets and the corresponding metrics in Table~\ref{tab:dataset-and-metrics}.
We evaluate different sampling strategies on 6 widely used academic benchmarks to measure the knowledge and reasoning abilities.
Here, we report the macro-averaged score of ARC-e, ARC-c, and BoolQ as the QA task performance.
Besides, open-ended user queries (e.g. creative writing) are more common in real scenarios, so we also evaluate methods on MT-Bench~\cite{Zheng2023mtbench}, which is aligned with human preferences.

\begin{table*}[]
    \centering
    \resizebox{\textwidth}{!}{
    \begin{tabular}{cccc p{7cm}}
    \toprule
        Dataset & \#Tasks & \#Few-shots & Metric & Introduction \\
        \midrule
        MMLU~\cite{hendryckstest2021mmlu} & 57 & 5 & Macro-averaged Accuracy & Multiple choice problems with a wide range of subjects, e.g. geography, history, etc. \\
        BBH~\cite{suzgun2022bbh} & 13 & 3 & Macro-averaged Exact Match & Reasoning over abstract reasoning tasks, e.g. logical expressions, causal judgement, etc. \\
        GSM8K~\cite{cobbe2021gsm8k} & 1 & 8 & Macro-averaged Exact Match & Grade school math problems with basic arithmetic operations (+-×÷). \\
        MBPP~\cite{austin2021mbpp} & 1 & 0 & Pass@1 & Generating Python function codes to pass test cases. \\
        ARC-e~\cite{clark2018arc} & 1 & 0 & Normalized Accuracy & Multiple-choice grade school level science question answering. \\
        ARC-c~\cite{clark2018arc} & 1 & 0 & Normalized Accuracy & Similar to ARC-e with challenging question answering pairs selected. \\
        BoolQ~\cite{clark2019boolq} & 1 & 0 & Accuracy & Given a passage and a question about world knowledge, answer YES or NO. \\
        MT-Bench~\cite{Zheng2023mtbench} & 8 & 0 & Subjective Score & Given a prompt and a generated response, using GPT-4~\cite{openai_introducing_chatgpt} to give scores from 1 to 10. \\
        \bottomrule
    \end{tabular}
    }
    \caption{Datasets and metrics for evaluations.}
    \label{tab:dataset-and-metrics}
\end{table*}

\subsection{Final Sampling Weights}
The final sampling weights of the proposed \textbf{Dynamic} method across MoE models are shown in Table~\ref{tab:final-sampling-weights}.
We find the two models show different preferences of instruction tuning datasets.
MoLM prefers ShareGPT while LLaMA-MoE prefers Math-Instruct.
This indicates that unified pre-defined sampling weights may not be suitable for all models, and we should devise sampling weights carefully according to their states.

\subsection{Performance Comparison with the Publicly Available SFT Model}

\begin{table*}[]
    \centering
    \begin{tabular}{lcccccc}
        \toprule
        Model & MMLU & BBH & GSM8K & MBPP & QA & MT-Bench \\
        \midrule
        w/o IT & 27.98  & 29.67  & 4.63   & 5.12   & 57.45  & -     \\
        LLaMA-MoE-SFT & 25.53 & 28.84 & 2.81 & 7.31 & 57.95 & 4.72\\
        Uniform & 32.48 & 29.18 & 5.91 & 14.52 & 60.85 & 5.07 \\
        Dynamic & \textbf{33.07} & \textbf{30.77} & \textbf{11.90} & \textbf{16.88} & \textbf{61.28} & \textbf{5.22} \\
        \bottomrule
    \end{tabular}
    \caption{Performances comparison with publicly available LLaMA-MoE-SFT.}
    \label{tab:public-sft-comparison}
\end{table*}
We provide the performance comparisons with publicly available SFT models in Table~\ref{tab:public-sft-comparison}.
Since MoLM does not have corresponding SFT versions of models, we present the performance comparisons between LLaMA-MoE-SFT~\cite{llama-moe-2023} and our fine-tuned LLaMA-MoE models, where these models are fine-tuned on the same foundation model.
Since LLaMA-MoE-SFT is only fine-tuned on a single dataset (ShareGPT), we find the simple Uniform baseline surpasses the public SFT model with large improvements, demonstrating the power of utilizing multiple instruction tuning datasets.
Besides, our proposed \textbf{Dynamic} outperforms \textbf{Uniform} with large margins, showing the effectiveness of dynamic sampling.

\subsection{Detailed Results of MT-Bench \& BBH}

Table~\ref{tab:multi-turn} shows the detailed multi-turn results on MT-Bench.
For better comparison the \textbf{Dynamic} effect on different tasks, we provide the detailed results on BBH subtasks in Table~\ref{tab:bbh-res}.

\begin{table*}[t]
    \centering
    \begin{tabular}{lcccc}
    \toprule
    \textbf{Model} & \textbf{ShareGPT} & \textbf{OpenOrca} & \textbf{Math-Instruct} & \textbf{Code Instructions} \\
    \midrule
    MoLM 700M-4E & 28.41 & 23.51 & 23.45 & 24.63 \\
    LLaMA-MoE 3.5B-2E & 17.98 & 21.49 & 32.02 & 28.51 \\
    \bottomrule
    \end{tabular}
    \caption{
Final sampling weights of \textbf{Dynamic} (\%).
The summation may not equal to exact 100\% due to digit rounding.
We find the final static weights of different models have many variations.
MoLM prefers to accept more ShareGPT, while LLaMA-MoE samples more Math-Instruct.
}
    \label{tab:final-sampling-weights}
\end{table*}

\begin{table*}[t]
    \centering
    \begin{tabular}{ccccccc}
    \toprule
    \multirow{2}{*}{\textbf{Rounds}} & \multicolumn{3}{c}{\textbf{MoLM}} & \multicolumn{3}{c}{\textbf{LLaMA-MoE}} \\
    & \textbf{DataSize} & \textbf{Uniform} & \textbf{Dynamic} & \textbf{DataSize} & \textbf{Uniform} & \textbf{Dynamic} \\
    \cmidrule(lr){1-1} \cmidrule(lr){2-4} \cmidrule(lr){5-7}
    1st & 2.81 & 2.98 & 3.10 & 5.52 & 5.78 & 5.96 \\
    2nd & 2.36 & 2.28 & 2.36 & 4.10 & 4.36 & 4.48 \\
    \rowcolor[gray]{.93}Overall & 2.59 & 2.63 & 2.73 & 4.81 & 5.07 & 5.22 \\
    \bottomrule
    \end{tabular}
    \caption{Detailed results on MT-Bench. Each question in MT-Bench has two turns of responses. Here we list the results of each turn.}
    \label{tab:multi-turn}
\end{table*}

\begin{table*}[t]
    \centering
    \resizebox{\textwidth}{!}{
    \begin{tabular}{ccccccc}
    \toprule
    \multirow{2}{*}{\textbf{Rounds}} & \multicolumn{3}{c}{\textbf{MoLM}} & \multicolumn{3}{c}{\textbf{LLaMA-MoE}}\\
    & \textbf{DataSize} & \textbf{Uniform} & \textbf{Dynamic} & \textbf{DataSize} & \textbf{Uniform} & \textbf{Dynamic} \\
    \cmidrule(lr){1-1} \cmidrule(lr){2-4} \cmidrule(lr){5-7}

Boolean Expressions & 53.20 & 54.40 & 55.20 & 49.20 & 47.20 & 46.80 \\
Causal Judgement & 36.90 & 52.94 & 51.87 & 52.94 & 52.41 & 50.80 \\
Date Understanding & 20.80 & 18.40 & 19.20 & 24.40 & 29.60 & 36.80 \\
Disambiguation Qa & 38.00 & 38.80 & 38.80 & 30.80 & 31.60 & 28.00 \\
Dyck Languages & 9.20 & 13.60 & 15.20 & 18.40 & 10.80 & 15.60 \\
Formal Fallacies & 37.60 & 39.60 & 21.60 & 49.20 & 53.20 & 52.40 \\
Geometric Shapes & 12.00 & 9.60 & 10.40 & 9.60 & 9.60 & 22.40 \\
Hyperbaton & 48.40 & 48.40 & 48.40 & 51.60 & 45.60 & 43.60 \\
Logical Deduction Five Objects & 8.40 & 21.20 & 22.80 & 18.40 & 22.80 & 20.00 \\
Logical Deduction Seven Objects & 10.00 & 17.20 & 14.40 & 15.60 & 15.60 & 14.40 \\
Logical Deduction Three Objects & 34.00 & 33.60 & 34.40 & 39.20 & 36.40 & 38.00 \\
Movie Recommendation & 14.80 & 22.40 & 19.60 & 41.60 & 22.40 & 26.00 \\
Multistep Arithmetic Two & 0.00 & 0.00 & 0.00 & 0.80 & 1.20 & 1.20 \\
Navigate & 32.40 & 42.40 & 46.40 & 50.80 & 56.40 & 50.80 \\
Object Counting & 14.80 & 16.80 & 13.20 & 33.20 & 33.60 & 38.40 \\
Penguins In A Table & 10.27 & 10.27 & 22.60 & 20.55 & 21.23 & 26.03 \\
Reasoning About Colored Objects & 1.60 & 7.60 & 13.20 & 7.60 & 14.00 & 21.60 \\
Ruin Names & 20.80 & 11.60 & 10.80 & 21.20 & 18.00 & 20.00 \\
Salient Translation Error Detection & 20.80 & 11.60 & 18.00 & 22.40 & 22.40 & 22.40 \\
Snarks & 48.31 & 51.69 & 52.25 & 55.62 & 46.63 & 60.67 \\
Sports Understanding & 46.00 & 54.00 & 54.40 & 56.00 & 58.40 & 57.60 \\
Temporal Sequences & 27.60 & 21.20 & 25.20 & 11.60 & 10.80 & 12.80 \\
Tracking Shuffled Objects Five Objects & 6.80 & 8.40 & 18.40 & 13.60 & 20.00 & 16.40 \\
Tracking Shuffled Objects Seven Objects & 7.20 & 14.00 & 14.00 & 12.80 & 15.20 & 14.80 \\
Tracking Shuffled Objects Three Objects & 33.20 & 32.80 & 36.00 & 33.60 & 33.60 & 32.00 \\
Web Of Lies & 51.20 & 50.40 & 49.60 & 49.60 & 51.60 & 53.60 \\
Word Sorting & 2.00 & 1.20 & 2.00 & 5.20 & 7.60 & 7.60 \\
\rowcolor[gray]{.93}Average & 23.94 & 26.08 & 26.96 & 29.46 & 29.18 & 30.77 \\
    \bottomrule
    \end{tabular}
    }
    \caption{Detailed results on different subtasks of BBH.}
    \label{tab:bbh-res}
\end{table*}

\end{document}